\documentclass[11pt,a4paper]{article}
\usepackage[hyperref]{emnlp2020}
\usepackage{graphicx}
\usepackage{times}
\usepackage{refcount}
\usepackage{latexsym}

\usepackage{microtype}

\aclfinalcopy 

\title{Lite Training Strategies for Portuguese-English \\  and English-Portuguese  Translation}

\author{Alexandre Lopes$^{1}$ \\
  \And
  Rodrigo Nogueira$^{2,3,4}$ \\
  \And
  Roberto Lotufo$^{2,3}$ \\
  \And
  Helio Pedrini$^{1}$ \\
  \AND
  \normalfont$^{1}$Institute of Computing, University of Campinas, Brazil \\
  $^{2}$School of Electrical and Computer Engineering, University of Campinas, Brazil \\
  $^{3}$NeuralMind Inteligência Artificial, Brazil \\
  $^{4}$David R. Cheriton School of Computer Science, University of Waterloo, Canada
  }

\date{August 2020}

\begin{document}

\maketitle

\begin{abstract}
Despite the widespread adoption of deep learning for machine translation, it is still expensive to develop high-quality translation models. In this work, we investigate the use of pre-trained models, such as T5 \cite{raffel2019exploring} for Portuguese-English and English-Portuguese translation tasks using low-cost hardware. We explore the use of Portuguese and English pre-trained language models and propose an adaptation of the English tokenizer to represent Portuguese characters, such as diaeresis, acute and grave accents. We compare our models to the Google Translate API and MarianMT on a subset of the ParaCrawl dataset, as well as to the winning submission to the WMT19 Biomedical Translation Shared Task. We also describe our submission to the WMT20 Biomedical Translation Shared Task. Our results show that our models have a competitive performance to state-of-the-art models while being trained on modest hardware (a single 8GB gaming GPU for nine days). Our data, models and code are available at \url{https://github.com/unicamp-dl/Lite-T5-Translation}.
\end{abstract}

\section{Introduction}

With the advent of deep neural networks, results in machine translation have recently improved over classical statistical strategies~\cite{wu2016google,artetxe-etal-2018-unsupervised}. For instance, in the Third and Fourth Conference on Machine Translation (WMT18~\cite{edunov-etal-2018-understanding} and WMT19~\cite{ng-etal-2019-facebook}), the top-performing systems for English-German and German-English competitions were based on transformers~\cite{vaswani2017attention}.

Transformer models are state-of-the-art architectures for MT tasks and are capable of translating the same word to different words based on the context. For instance, the word 'bank' in Portuguese can be translated to 'bench' or 'bank' depending on the context. 

\begin{figure*}[!htb]
\centering
\includegraphics[width=13cm]{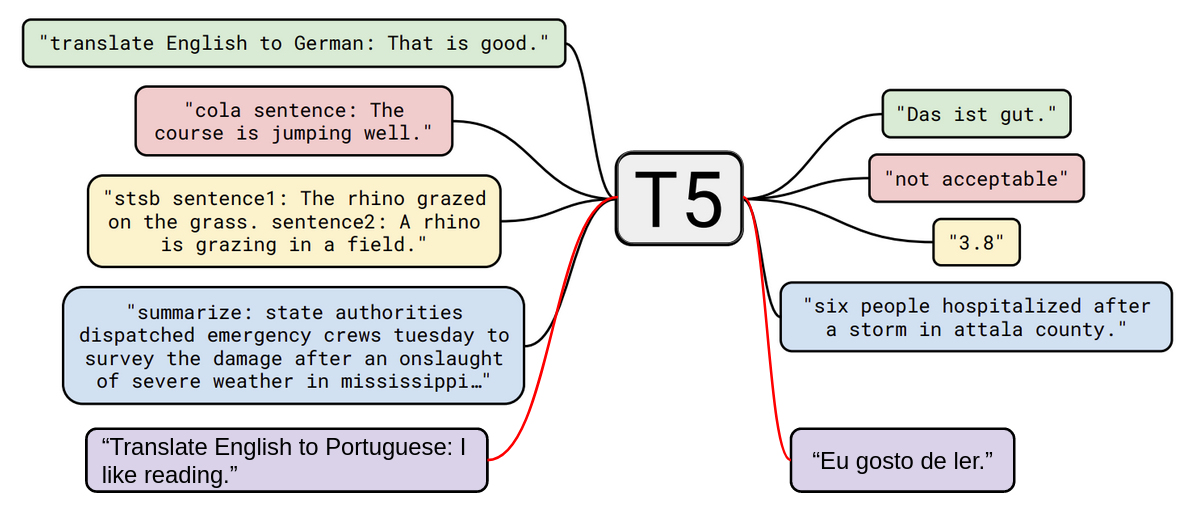}
\caption{\label{fig:t5}The text-to-text framework used by T5. The purple boxes and red connections represent the task used in this work. Figure adapted from~\cite{raffel2019exploring}.}
\end{figure*}

This work explores translation strategies using language models pre-trained on Portuguese and English corpora. More specifically, we investigate the use of \textbf{T}ext-to-\textbf{T}ext \textbf{T}ransfer \textbf{T}ransformer (T5) pre-trained model for these tasks. An illustration of T5 for the English-Portuguese translation task is shown in Figure~\ref{fig:t5}. The main contributions of this work are:

\begin{itemize}
\item We show that it is possible to train translation models that are competitive with the state of the art using few computational resources. We trained our models on a gaming desktop with an Nvidia RTX2070 GPU, i5 CPU, and 32GB RAM. In comparison, the winning submission of the WMT19 Biomedical Translation Shared Task used four NVIDIA V100 GPUs, each being approximately ten times more expensive than an RTX2070.

\item We created and made public ParaCrawl 99k, a dataset of 99k sentence pairs extracted from ParaCrawl's English-Portuguese parallel corpus\footnote{\url{https://paracrawl.eu/}}. This large test corpus allows researchers to evaluate their models on a general-domain translation task.

\item We evaluated Google Translate on ParaCrawl 99k, allowing other researchers to compare their results to a high-quality commercial system.

\item We developed an adaptation for the English pre-trained tokenizer and achieved better results on English-Portuguese translation tasks than using the tokenizer without any changes. This allows us to efficiently adapt language models to a vocabulary that was not seen during pre-training.

\end{itemize}

\section{\label{sec2}Related Work}

Two widely adopted types of MT systems are Statistical Machine Translation (SMT) systems and Neural Machine Translation (NMT) systems~\cite{sutskever2014sequence,bahdanau2014neural}. The first one relies on statistical techniques to perform translation, such as counting the number of times a word occurs in the context of other words. A popular example of such system is Moses~\cite{koehn-etal-2007-moses}.

The winning system of WMT'19 Biomedical competition for en-pt and pt-en translation tasks~\cite{soares-krallinger-2019-bsc} is an NMT system. They used OpenNMT-py to train a transformer model on seven parallel corpora. However, differently from our models, their model was trained from scratch.

Recent works~\cite{peters-etal-2018-deep, devlin2018bert} have shown the advantages of using pre-trained models for tasks such as question-answering and text classification. The intuition is to allow the network to use information from pre-training language representations to increase the performance on specific tasks.

\citet{edunov-etal-2019-pre} evaluated the use of a pre-trained encoder-decoder model for translation. Both encoder and decoder weights were tied, but they were pre-trained on different languages. This is an expensive strategy for techniques that use a trainable tokenizer, such as SentencePiece~\cite{kudo-richardson-2018-sentencepiece}, because it is necessary to re-train the entire model if the vocabulary changes, as new token embeddings need to be learned.

Many commercial systems, such as Google Translate (GT) and Amazon Translate (AT), have an excellent performance on MT, but they are expensive if one needs to translate vast amounts of text. For example, we estimate that it would cost 50,000 USD to translate the 20 million sentences of ParaCrawl using GT. Unfortunately, no commercial system that we are aware of provides metric scores on public datasets that would allow us to compare their systems to ours.

\section{\label{sec3}Methods}

We proposed two main strategies for translating: using a T5 model pre-trained on a Portuguese corpus and adapting the original T5 tokenizer to work with Portuguese texts. 

\subsection{Pre-trained Language Model}

We evaluated four different scenarios: English-Portuguese translation with T5 pre-trained in a Portuguese corpus; English-Portuguese translation with T5 pre-trained in an English corpus; Portuguese-English translation with T5 pre-trained in an English corpus; and Portuguese-English translation with T5 pre-trained in a Portuguese corpus.  

These variations allow us to evaluate how the language used during pre-training affects the translation's performance. 

\subsection{\label{sec31}Adaptation of the English tokenizer to Portuguese}

Here we investigate if we can adapt to the English-Portuguese translation task a model already pre-trained on languages other than Portuguese.

We observe that using a non-Portuguese tokenizer can cause translation problems, since some Portuguese characters cannot be represented, such as letters with the tilde accent (e.g. 'ã'). To fix this issue, we propose an adaptation of the original T5 tokenizer using a pre-processing and post-processing strategy.
The tokenizer's adaptation allows the tokenizer to represent all possible characters in the Portuguese language. 

We can divide this adaptation into two stages: Token Completion and Word Regrouping. The first stage allows the use of Portuguese special characters, such as accented vowels, whereas the second stage merges these extra tokens back to form correct words.

\subsubsection{Token Completion Stage}

In this step, we start adding to the tokenizer Portuguese accented vowels that were not present in it. We ended up adding fourteen of those characters, as well as the word 'não', which is the most common word in the ParaCrawl pt-en dataset.

A list of all added tokens is available in Table~\ref{table:t1}. The addition of these tokens allowed the model to learn and generate them in en-pt translation.

This is also an inexpensive method for increasing the number of words that can be represented, since only the embeddings of the new tokens have to be learned from scratch. The existing token embeddings, which represent the majority of the non-Portuguese tokens, were already learned during the pre-training phase and can be reused in the fine-tuning phase.  

We show in Table~\ref{table:t2} some encoding and decoding examples after adding tokens to the tokenizer.

\begin{table}[!htb]
\centering
\begin{tabular}{lclclclc}
\hline
ì ò Á Í Ó Ú í ú Â Ê Ã Õ ã õ não \\
\hline
\end{tabular}
\caption{\label{table:t1} List of tokens added to the T5 tokenizer by our adaption method. }
\end{table}

\begin{table}[!htb]
     \centering
        \begin{tabular}{lclclclc}
			\hline
			\hline
			\textbf{Tokenizer without additional Port. tokens}\\
			 original $\rightarrow$ after encoding/decoding \\
			\hline
			eu gosto de arroz $\rightarrow$ eu gosto de arroz \\
			eu não como $\rightarrow$ eu n ? o como \\
			indignação completa $\rightarrow$ indignaç ? o \\ completa \\
			\hline
			\hline
			\textbf{Tokenizer with additional Port. tokens}\\
			 original $\rightarrow$ after encoding/decoding \\
			\hline
			eu gosto de arroz $\rightarrow$ eu gosto de arroz \\
			eu não como $\rightarrow$ eu não como \\
			indignação completa $\rightarrow$ indignaç ã o \\ completa \\
			\hline
        \end{tabular}
    \caption{\label{table:t2} Comparing tokenizer results before and after adding the Portuguese tokens.}
\end{table}

\subsection{Word Regrouping Stage}

When adding tokens directly to the tokenizer, the HuggingFace's~\cite{Wolf2019HuggingFacesTS} SentencePiece implementation used in our work interprets the result as a new complete token, i.e., not part of a word. For example, the word 'pão' is broken into three different tokens 'p' 'ã' 'o'. This is fixed in a post-processing step called Word Regrouping. 
    
In this step, we regroup the added tokens of vowels with accents separated erroneously by the tokenizer. We find in the translated text all tokens added in the Token Completion step, and merge them with their neighboring words.

\begin{figure}[!htb]
\includegraphics[width=0.5\textwidth]{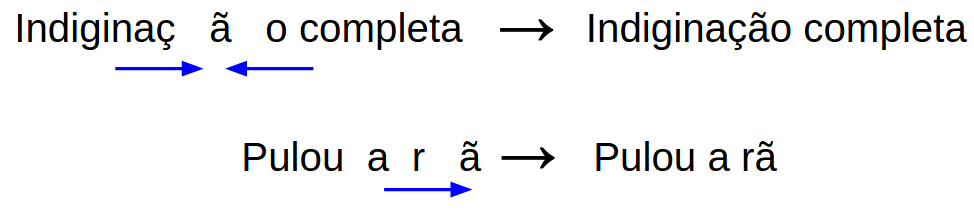}
\caption{\label{fig:m1}An example of separated tokens merged back into a single word. Our algorithm searches for an isolated special token (in this case, 'ã') and merges it with its neighbors. It can be merged at the beginning, middle, or end of a sentence.
}
\end{figure}

In Figure~\ref{fig:m1}, we illustrate our algorithm.
   
\vspace{-0.35cm}
\section{\label{sec4}Datasets}

We trained our models using six different datasets, and we evaluate our system on two datasets: the WMT19 Biomedical Translation Task dataset and a subset of 99,000 sentence pairs of the ParaCrawl dataset. We also present the results of our submission to the WMT20 Biomedical Translation Task competition.

\subsection{Training Datasets}

We have two different strategies for training our models depending on the test datasets. For the evaluation on the ParaCrawl dataset, we only trained the models on ParaCrawl data. ParaCrawl is a public parallel corpus of many European languages available online. It contains approximately 20M English-Portuguese sentence pairs. Due to our small computational budget, we randomly selected approximately 5M pairs for training.

For WMT19 and WMT20 Biomedical Translation Tasks, we train our models on the ParaCrawl dataset as well as on the following datasets, which are of the same domain as WMT's Biomedical data:

\begin{itemize}
\item EMEA Corpus~\cite{TIEDEMANN12.463}: A parallel corpus of European Medicines Agency documents.

\item CAPES Parallel Dataset~\cite{soares2018parallel}: A parallel corpus of theses and dissertations abstracts collected from the CAPES website.

\item Scielo Dataset~\cite{soares2018large}: A parallel corpus of scientific articles collected from SciELO.

\item JRC-Acquis~\cite{steinberger-etal-2006-jrc}: A parallel corpus of European Union (EU) documents in all official EU languages.

\item Biomedical Domain Parallel Corpora~\cite{neveol-etal-2018-parallel}: A repository of the challenge that contains links to different parallel corpora. We used the Medline, Scielo, and ReBEC training datasets.

\end{itemize}

Besides being of the same domain of WMT's Biomedical task, an advantage of these datasets over ParaCrawl is that they are in Brazilian Portuguese, such as most of WMT's Biomedical data.
The number of sentence pairs used for training from each dataset is shown in Table~\ref{table:t7}.

\begin{table}[!htb]
\setlength{\tabcolsep}{3.0mm}
\centering
\begin{tabular}{lr}
\hline
\textbf{Corpus} & \textbf{Sent. Pairs} \\
\hline
EMEA & 1,082,144 \\
CAPES & 1,157,610  \\
Scielo & 2,828,916 \\ 
JRC-Acquis & 1,236,846 \\ 
Biomedical Domain Corpora & 331,937  \\ 
\hline
\textbf{Total} & \textbf{6,637,453} \\
\hline
\end{tabular}
\caption{\label{table:t7} Number of sentence pairs of each domain-specific dataset used to train our models for the WMT19 and WMT20 Biomedical tasks.}
\end{table}

\subsection{Testing Datasets}

We created a general-domain test set from the ParaCrawl dataset. We begin by randomly selecting 128,000 sentence pairs from its 20M pairs. ParaCrawl is originally deduplicated, but similar sentences still might exist in our split of the training and test sets. Thus, we apply as stricter deduplication process to increase the quality of our test set. To increase the speed in verifying similarity of sentence pairs, we used MinHash and Locality-Sensitive Hashing (LSH)~\cite{rajaraman2011mining} to compare sentences of training and test datasets. We set a Jaccard similarity threshold to 0.7, i.e., all sentences with similarity greater than 0.7 were discarded from the test set. LSH found 28,913 sentences in the test set with a similarity score above 0.7 of sentences in the training set. The final test set ended up having 99,087 sentence pairs, which we called ParaCrawl 99k test set. 
This dataset and its corresponding translations using GT are available in our Github.

We also evaluated our system on the WMT19 Biomedical Shared Task test set. This is a dataset composed of approximately 500 parallel sentences of Medline abstracts. 

Finally, we submitted our results to the WMT20 Biomedical Shared Task competition. The WMT20 test set has 544 parallel sentences for the English-Portuguese translation task and 498 sentences for the Portuguese to English task.

\section{\label{sec5}Experiments}

We conducted several experiments using different configurations of T5. We divided the experiments into two groups: model hyperparameter optimization and different pre-training studies. All experiments were performed on a desktop computer with an Nvidia 8GB RTX 2070 Super, 32 Gb RAM memory, and a 4-core Intel processor running on Ubuntu 18.04. We used PyTorch~\cite{paszke2017automatic}, HuggingFace Transformer, and Pytorch-Lightning~\cite{falcon2019pytorch} frameworks to train and evaluate our models.

\subsection{Model Hyperparameter Optimization}

We tuned the hyperparameters using the original T5 checkpoint available in the HuggingFace library. This model was pre-trained on a corpus whose majority of documents were in English with a small proportion of German, French, and Romanian documents. We first conducted a small training using 100k sentence pairs and evaluated on another 50k sentence pairs to determine some hyperparameters of the T5 model, such as batch size and maximum length of tokens in the source and target sentences. We also evaluated the optimizer and found the best convergence with the AdamW Optimizer~\cite{loshchilov2017decoupled}. All hyperparameters used are in Table~\ref{table:t4}. With this configuration, we evaluated the performance of adding Portuguese-only characters to the tokenizer in comparison to using the original T5 tokenizer. The results are available in Table~\ref{table:t3}. Our proposed tokenizer adaption resulted in an improvement of almost 5 BLEU points over the original tokenizer in the en-pt translation task.

\begin{table}[!htb]
\setlength{\tabcolsep}{3.0mm}
\centering
\begin{tabular}{lr}
\hline
\textbf{Hyperparameters} & \textbf{Values} \\
\hline
Batch Size & 256 \\
Source Sequence Length (SSL)& 96 \\
Target Sequence Length (TSL) & 160 \\
Learning Rate & $5\cdot10^{-3}$ \\
eps & $1\cdot10^{-5}$ \\
\hline
\end{tabular}
\caption{\label{table:t4} Hyperparameters used for training the models.}
\end{table}

\begin{table}
\setlength{\tabcolsep}{3.0mm}
\centering
\begin{tabular}{ll}
\hline
\textbf{Translation Type} & \textbf{SacreBLEU} \\
\hline
Original T5 tokenizer & 31.15 \\
\qquad + Portuguese characters & 35.95 (+4.8)\\ 
\hline
\end{tabular}
\caption{\label{table:t3} Effects in performance of using our adaption of the original T5 tokenizer to the English-Portuguese translation task. Numbers are from ParaCrawl's 99k en-pt test set.}
\end{table}

After finding these hyperparameters, we analyzed the trade-off between model sizes in a subset of the ParaCrawl dataset of 1M sentence pairs and evaluated them in a 150k sentence subset. We did not use any sentence from the test set. The results of this analysis are reported in Table~\ref{table:t5}. We trained the T5-small and T5-base models with different epoch sizes. Training 3 epochs of T5-small takes almost the same time as training one epoch with a T5-base model.

The performance would possibly increase if we used large models such as T5-large, T5-3B, or T5-11B. However, we could fit only the T5-base model in our 8GB GPU. We used batch accumulation to achieve batches of size 256 as the T5-small can only handle batch size 4 in 8GB. Thus, one of the contributions of this work is to show that it is possible to train translation models that are close to the state of the art on a relatively inexpensive hardware.

All experiments in the following sections using Tokenizer's Adaptation Steps (\ref{sec31}) were performed using the best pre-processing and post-processing strategies presented in Table~\ref{table:t5}.

\begin{table}[!htb]
\setlength{\tabcolsep}{1.2mm}
\centering
\begin{tabular}{lr}
\hline
\textbf{Translation Type} & \textbf{Sacre}\\{} &\textbf{BLEU} \\
\hline
Adding Top 25 words in Port. \\+ T5-small + 3 epochs & 43.03 \\
Adding tokens of Table~\ref{table:t1} in Port. \\+ T5-small + 3 epochs & 43.48 \\ 
Adding tokens of Table~\ref{table:t1} in Port. \\+ T5-base + 1 epoch & \textbf{44.52} \\ 

\hline
\end{tabular}
\caption{\label{table:t5} Effects in performance of different strategies for adapting the original T5 tokenizer to Portuguese. Numbers are from  our dev set of ParaCrawl.}
\end{table}

\subsubsection{Pre-training Studies}

We also evaluated the effects of pre-training the model in a corpus of the same language of the target language. The intuition here is that it would be easier for the model to learn the target language than having previous knowledge of the source language. Since the tokenizer mainly has tokens of one of the two languages, it is better to have a smaller quantity of tokens to learn. This is because, if the SentencePiece tokenizer does not have the word in its vocabulary, it will use subtokens to form the original word. For example, the sentence 'They like to drink coconut water' is represented by six tokens in English SentencePiece and thirteen tokens in Portuguese SentencePiece. We are not evaluating here the possibility to train the pre-training model from scratch with both languages together, as it is not possible with our modest hardware setup.

For the Portuguese pre-trained model, we used PTT5-base model~\cite{ptt5_2020} with Portuguese tokenizer. PTT5 was pre-trained on BrWAC, a large corpus of Brazilian Portuguese webpages. PTT5 started training using T5's official published weights as initial weights, so it also uses English learning to its model. For the English pre-trained model, we used the Huggingface implementation of T5 with its default tokenizer, which is based on SentencePiece.

In Table~\ref{table:t6}, we compare both models with Google Translate in the ParaCrawl 99k test set. Both models perform similarly in the Portuguese-English translation task, but the Portuguese pre-trained model gives a better result than the English pre-trained model in the English-Portuguese translation task. We are on par with Google Translate on en-pt, but a few BLEU points below on pt-en.

\begin{table}[!htb]
\setlength{\tabcolsep}{1.0mm}
\centering
\begin{tabular}{lrr}
\hline
\textbf{} & \textbf{pt-en} & \textbf{en-pt}\\
\hline
Google Translate API & 51.20 & 45.17 \\
\hline
Ours - English pre-training & 46.49 & 44.56 \\ 
Ours - Portuguese pre-training & 46.35 & 45.44 \\ 
\hline

\end{tabular}
\caption{\label{table:t6} SacreBLEU comparison between GT and our approach in Paracrawl 99k test set.}
\end{table}

\section{\label{sec6}WMT19 and WMT20 Results}

We now evaluate our models on the WMT19 Biomedical Translation Task and our show the results of our official submission to the WMT20 Biomedical Translation Task.

In Table~\ref{table:t8}, we show WMT19 results of our models as well as the winning submission of WMT19 Biomedical tasks~\cite{soares2019bsc} and the MarianMT~\cite{junczys2018marian} implementation available on the HuggingFace's Transformer Library.\footnote{\url{https://huggingface.co/transformers/model_doc/marian.html}} Models pre-trained on Portuguese obtained the best performance in both translation tasks. Notably, we achieved an improvement of +6.31 BLEU points in the English to Portuguese translation task by using the Portuguese pre-trained model and +9.75 with an increase of target and source sequence lengths. We also obtained an improvement of +0.62 in the Portuguese to English translation task using the Portuguese pre-trained model and +2.27 when increasing target and source sequence lengths.

We believe that the improvement of Portuguese pre-training models is associated with PTT5's training strategy that uses English pre-trained weights as initial weights. The intuition is that PTT5 carries information from the English model too. 

\begin{table}[!htb]
\setlength{\tabcolsep}{1.4mm}
\centering
\begin{tabular}{lrr}
\hline
\textbf{} & \textbf{pt-en} & \textbf{en-pt}\\
\hline
MarianMT & 27.91 & 47.44 \\
BSC \\ \cite{soares2019bsc} & 39.90 & 48.18 \\
\hline
Ours - English pre-training & 45.89 & 39.31 \\
\hline
Ours - Portuguese pre-training & 46.51 & 45.62 \\ 
\qquad + TSL=256 and SSL=256 & \textbf{$-$} & \textbf{49.06} \\ 
\qquad + TSL=140 and SSL=160 & \textbf{48.16} & \textbf{$-$} \\ 

\hline
\end{tabular}
\caption{\label{table:t8} BLEU scores on the test set of WMT19 Biomedical Shared Task. Portuguese pre-training was tested in three different scenarios: one with default hyperparameters available in Table~\ref{table:t8} and two with different Target Sequence Length (TSL) and Source SEquence Length (SSL).}
\end{table}

The results for WMT20's challenge are in Table~\ref{table:t9}. Our submission is 2.17 BLEU points below the winning team in Portuguese-English, but it is 4.48 BLEU points higher than the baseline. For the English-Portuguese task, our results are below the baseline. That can be attributed to not using the Portuguese pre-trained model, which was not available at the time of our submission. As noted above, we achieved a large improvement on WMT19 when we switched from the English to the Portuguese pre-trained model. Therefore, we assume that a Portuguese pre-trained model would obtain superior results to the baseline on WMT20.

\begin{table}[!htb]
\setlength{\tabcolsep}{4.0mm}
\centering
\begin{tabular}{lll}
\hline
\textbf{Team Names} & \textbf{pt-en} & \textbf{en-pt}\\
\hline
Sheffield & 48.16 & 44.57 \\
\textbf{Unicamp\_DL}  & 45.99 & 38.08$^{\star}$ \\
baseline  & 41.51 & 39.77 \\ 
\hline
\end{tabular}
\caption{\label{table:t9} BLEU scores on WMT20's automatic evaluation. $^{\star}$Since the Portuguese T5 model was not available at the time of our submission, we used the original (English) T5. Hence, results for en-pt can now be improved by switching to the Portuguese pre-trained model.}
\end{table}
\vspace{-0.5cm}
\section{\label{sec7}Conclusions and Future Work}

We show that it is possible to develop English-Portuguese translation models close to the state of the art using modest hardware. Despite not reaching the same level of performance of Google Translate on pt-en, the fact that our system was developed mostly by the first author on its personal computer shows that implementing high-quality machine translation systems has become possible for anyone, including small companies and research labs.

We also presented our submission strategies for the WMT20 Biomedical Translation Shared Task using a T5 model. We show that a simple adaption of the original T5 tokenizer to the Portuguese language largely improves the translation quality and does not require any further pre-training, which is expensive. However, we achieve the best results with models pre-trained on Portuguese.

As directions for future work, we plan to experiment with larger models and models pre-trained in both Portuguese and English languages simultaneously, as recent work showed that this a successful strategy~\cite{wu2016google,arivazhagan2019massively}. We believe that we could improve the translation results with larger and more complex models~\cite{lepikhin2020gshard}.

\section{Acknowledgements}

We thank CNPq research funding, process number 310828/2018-0.

\bibliographystyle{acl_natbib}
\bibliography{emnlp2020}

\end{document}